\documentclass{article}
\usepackage{spconf,amsmath,graphicx,hyperref}
\usepackage{stfloats} 
\usepackage{multirow}
\usepackage{booktabs}

\title{DSTCS: Dual-Student Teacher Framework with Segment Anything Model for Semi-Supervised Pubic Symphysis-Fetal Head Segmentation}
\name{Yalin Luo \qquad Shun Long \qquad Huijin Wang \qquad Jieyun Bai$^{\star}$\thanks{$^{\star}$Corresponding author. The work is funded by the Guangzhou Science and Technology Program (No. 2023B03J1297).}}

\address{School of Information Science and Technology, Jinan University, Guangzhou, China}
	
\begin{document}
\topmargin=0mm
\ninept
\maketitle
\begin{abstract}
Segmentation of pubic symphysis and fetal head (PSFH) is a critical procedure in intrapartum monitoring, essential for evaluating labor progression and identifying potential delivery complications. However, achieving accurate segmentation remains a significant challenge due to class imbalance, ambiguous boundaries, and noise interference in ultrasound images, compounded by a scarcity of high-quality annotated data. Current research on PSFH segmentation predominantly relies on CNN and Transformer architectures, leaving the potential of more powerful models underexplored. We propose a Dual-Student and Teacher combining CNN and SAM (DSTCS) framework, which integrates the powerful Segment Anything Model (SAM) into a dual student-teacher architecture. A cooperative learning mechanism between CNN and SAM branches significantly improves segmentation accuracy. The proposed scheme also incorporates a specialized data augmentation strategy optimized for boundary processing and a novel loss function.
Extensive experiments on the MICCAI 2023 and 2024 PSFH segmentation benchmarks demonstrate that our method exhibits superior robustness and significantly outperforms existing techniques, providing a reliable segmentation tool for clinical practice.
\end{abstract}

\begin{keywords}
Segment Anything Model, Pubic Symphysis-Fetal Head Segmentation, Semi-Supervised Learning
\end{keywords}

\section{Introduction}
Segmentation of pubic symphysis and fetal head (PSFH) in intrapartum ultrasound imaging has to be both fast and robust to accommodate the dynamic and variable nature of the birth canal—a significant departure from the relatively stable conditions of prenatal ultrasound. Accurate segmentation is expected to not only address extreme class imbalance caused by the coexistence of multiple targets (e.g., fetal head and pubic symphysis) but also enable precise quantification of key parameters \cite{angeli2022automatic, bai2024new, chen2024psfhs, chen2024fetal}, assessment of fetal position, and mitigation of delivery risks. These demands place higher requirements on the generalization capability of segmentation algorithms.
Although substantial progress has been made in medical image segmentation in recent years, significant limitations remain in PSFH segmentation for intrapartum ultrasound. Traditional methods such as randomized Hough transform \cite{lu2005automated}, random forests \cite{van2018automated} the two-step detection method \cite{wang2014detection}, and the learning-based framework \cite{li2017automatic} are limited by inherent challenges in ultrasound imaging, including noise, low contrast, and blurred boundaries, resulting in limited generalization and complex processing pipelines. With the rise of deep learning, fully supervised approaches like U-Net \cite{ronneberger2015u} and its variants \cite{lu2022multitask, chen2024direction, bai2022framework} have improved segmentation accuracy. But they rely heavily on large annotated datasets, making clinical adoption costly. To reduce annotation requirements, semi-supervised learning (SSL) methods have been widely explored, such as Mean Teacher \cite{tarvainen2017mean} Interpolation Consistency \cite{verma2022interpolation}, Deep Adversarial Network \cite{zhang2017deep}, Deep Co-Training \cite{qiao2018deep}, cross pseudo-supervision \cite{chen2021semi}, and Cross Teaching between CNN and Transformer \cite{luo2022semi}. However, most established approaches are based solely on CNN or Transformer architectures and still underperform when handling irregular boundaries and dynamic deformations in ultrasound images. Furthermore, due to their lack of effective modeling of global anatomical constraints and dedicated optimization strategies for boundary ambiguity and noise sensitivity, current research has not fully exploited the potential of large visual models (e.g., Segment Anything Model, SAM) in ultrasound imaging.

To address these challenges, we propose DSTCS which incorporates the following key innovations:
(1) Dual Student-Teacher Framework: A dual-branch heterogeneous architecture integrating CNN and SAM adapters mitigates noise sensitivity and instability in pseudo-labels. The CNN extracts local texture features, while SAM delivers zero-shot generalization and spatial priors. Through cross-validation of hard/soft pseudo-labels and classifier alignment, it effectively suppresses ultrasound artifact-induced label noise while enhancing consensus learning and anti-interference capabilities in complex scenarios.
(2) Edge-Patch In-situ Superposition: We adopt an EPIS strategy that prioritizes the preservation of anatomical boundary information during augmentation, in order to avoid the corruption of key regions common in conventional augmentation methods.
(3) Neighborhood Weighted Dice Loss: The loss incorporates spatial weighting to enhance the model’s geometric sensitivity to boundary regions between the pubic symphysis and fetal head, addressing both class imbalance and boundary ambiguity.

\section{Method}

\begin{figure*}[htb]
	\centering
	\includegraphics[width=0.91\linewidth]{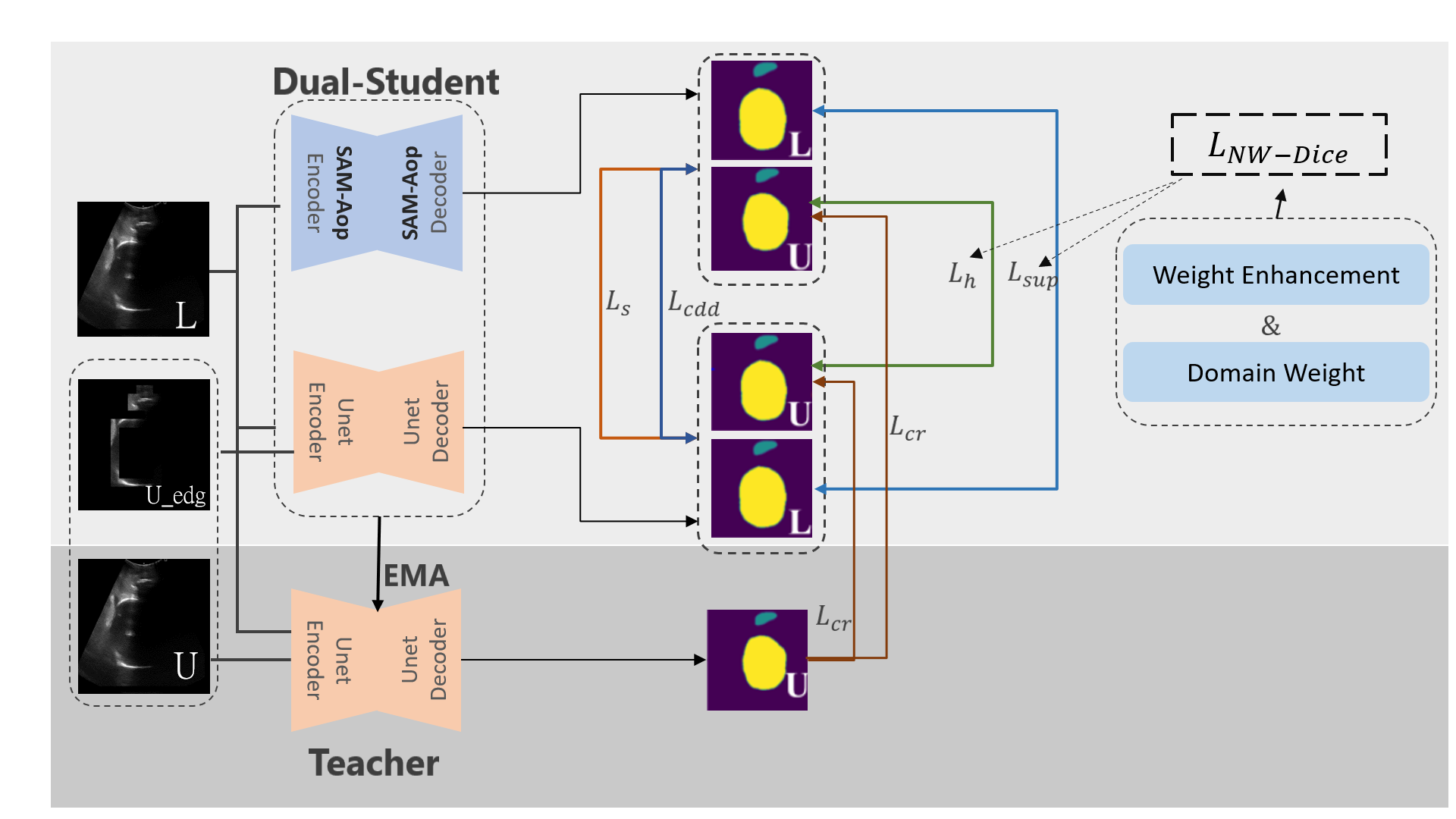}
	\caption{An overview of the DSTCS architecture, where the black and grey parts are the loss functions for student1 and student2}
	\label{fig:architecture}
\end{figure*}

For general semi-supervised learning, the training set is composed of a labeled dataset $\mathcal{D}_N^l$ with $N$ labeled images and an unlabeled dataset $\mathcal{D}_M^u$ with $M$ ($M \gg N$) raw images. It is normally denoted as $\mathcal{D}_{N+M} = \mathcal{D}_N^l \cup \mathcal{D}_M^u$. Given an image $X$, its ground truth Y is available if $X \in \mathcal{D}_N^l$, but not so if $X \in \mathcal{D}_M^u$. $P_{s1}$ and $P_{s2}$ are the probability outputs derived from student1 and student2 respectively. $P_{s1}^\ast$ is student1's soft pseudo labels.

DSTCS incorporates a SAM model to enhance the original dual student-teacher framework \cite{jiang2024intrapartum}. Specifically, the dual-student model consists of a CNN-based UNet (namely student1) and a SAM model called AoP-SAM\cite{zhou2025segment} (namely student2), while the single-teacher model employs the same CNN-based UNet. The design of AoP-SAM achieves a balance among computational efficiency, architectural synergy, and domain adaptability. Internally, it employs a parallel CNN branch (with multi-scale modules) that works synergistically with a ViT encoder integrated with LoRA-inspired feature adapters. Through a cross-branch attention mechanism, deep feature fusion is achieved, significantly enhancing the model's ability to capture multi-scale medical image features. Simultaneously, by unfreezing the prompt encoder and mask decoder, the model learns to generate internal guidance signals. This mechanism not only substantially reduces GPU memory requirements but also enables efficient and targeted adaptation to the ultrasound domain.

DSTCS tackles semi-supervised image segmentation from five various aspects: supervised learning ($L_{\mathrm{sup}}$), cross-supervision with hard pseudo labels ($L_h$), consistency learning with soft pseudo labels ($L_s$), minimization of classifier determinacy discrepancy ($L_{\mathrm{cdd}}$) and consistency regularization constraints from the teacher model ($L_{\mathrm{cr}}$). Therefore, the overall training loss function for student1 or student2 can be defined as:
\begin{equation}
L_{\mathrm{total}} = L_{\mathrm{sup}} + \alpha L_h + \beta L_s + \gamma L_{\mathrm{cdd}} + \mu L_{\mathrm{cr}}
\end{equation}
where $\alpha$, $\beta$, $\gamma$ and $\mu$ are trade-off weights, which were set $\alpha = 0.5$, $\beta = 1.0$, $\gamma = 3.0$, and $\mu = 0.1$ \cite{wang2022cnn} respectively in the proposed cooperative training process.
Note: $L_{\mathrm{cdd}}$ is a common part of both students.

{\bf Supervised learning ($L_{\mathrm{sup}}$).} We train the student models with labeled data. Cross-entropy loss $L_{\mathrm{ce}}$ and Dice loss $L_{\mathrm{dice}}$ are used as follows:
\begin{equation}
L_{\mathrm{sup}} = \frac{1}{2}\sum_{X \in \mathcal{D}_N^l} \left( L_{\mathrm{ce}}(P_{s1}, Y) + L_{\mathrm{dice}}(P_{s1}, Y) \right)
\end{equation}

{\bf Cross-supervision with hard pseudo labels ($L_h$).} The predictions generated by CNN and SAM exhibit distinct characteristics, particularly at the output level. Based on these predictions, we utilize the $\mathrm{argmax}(\cdot)$ function to produce hard pseudo-labels for cross-supervision between the two peer networks. The cross-supervision loss for unlabeled data is defined as:
\begin{equation}
L_h = \sum_{X \in \mathcal{D}_M^u} L_{\mathrm{dice}}\left( P_{s1}, \mathrm{argmax}(P_{s2}) \right)
\end{equation}

{\bf Consistency learning with soft pseudo labels ($L_s$).} To further reduce the noise of hard pseudo labels and focus on unlabeled challenging regions, a sharpening function \cite{li2022collaborative, krizhevsky2017imagenet} is utilized to generate soft pseudo labels, which aims to decrease the prediction uncertainty of the models. Soft pseudo labels can be obtained as follows:
\begin{equation}
P_{s1}^\ast = \frac{{(P_{s2})}^{1/\tau}}{{(P_{s1})}^{1/\tau} + (1 - P_{s2})^{1/\tau}}
\end{equation}
where $\tau$ is a hyper-parameter to control the temperature of sharpening, and is set to 0.1 in our experiment \cite{wu2022mutual}. Consistency learning is performed between the probability output of one model and the soft pseudo label of the other. The final loss function of consistency learning with soft pseudo labels is defined as :
\begin{equation}
L_s = \sum_{X \in \mathcal{D}_{N+M}} E\left[ P_{s1}, P_{s1}^\ast \right]
\end{equation}
where $E$ is the Mean Squared Error (MSE) loss.

{\bf Classifier Determinacy Disparity ($L_{cdd}$).} Based on the architectural differences between CNN (excelling at local feature extraction) and SAM (strong in global context modeling), a dual-classifier prediction relevance matrix $A = {P}_{s1} {P}_{s2}^T$ \cite{li2021bi} is constructed. The sum of the diagonal elements of matrix $A$ represents prediction consistency, while the sum of the off-diagonal elements reflects prediction uncertainty. The loss function is defined as:
\begin{equation}
L_{cdd} = \sum_{{X} \in \mathcal{D}_{N+M}} \left[ \sum_{m,n=1}^{C} {A}^{m,n} - \sum_{m=1}^{C} A^{m,m} \right]
\end{equation}
where ${A}^{m,n}$ denotes the element in the $m$-th row and $n$-th column, and $C$ is the number of categories.

{\bf Consistency Regularization ($L_{\mathrm{cr}}$).} The teacher module aims to minimize discrepancies between the predictions of the dual-student networks and the teacher network under both data and network perturbations. The consistency loss between the output probabilities of them is defined as follows:
\begin{equation}
L_{\mathrm{cr}} = \sum_{X \in \mathcal{D}_{M}} \left[ E\left(f_t(X;\bar{\theta};\sigma'), f_{s1}(X;\varphi;\sigma)\right) \right]
\end{equation}
where $\sigma$ and $\sigma'$ represent different data perturbations and random dropout operations at the network layer, $\varphi$ and $\bar{\theta}$ represent the network parameter of student1 and teacher, $\bar{\theta}$ is updated via exponential moving average (EMA) from student2's network, and $E$ is the MSE loss.

{\bf Edge-Patch In-situ Superposition (EPIS).} The UNet branch innovatively incorporates EPIS strategy. EPIS divides the original image into a grid of $16 \times 16$ pixels, locates anatomical edge regions through label $Y$ to extract a set of edge local patches, and re-embeds them into the enhanced image $X'$. Specifically, it can be expressed as follows:
\begin{equation}
X' = \mathcal{R}(X^l, Y^l) + \mathcal{R}(X^u, \arg\max f_\theta(X^u))
\end{equation}
where $X'$ denotes the enhanced image, $\mathcal{R}$ represents the edge reorganization function, $l$ corresponds to labeled data, $u$ denotes unlabeled data, and $f_\theta$ refers to the model prediction function along with its network parameters. This mechanism significantly enhances the model's capability to recognize edge features.

{\bf Neighborhood Weighted Dice Loss.} ($\mathcal{L}_{\text{NW-Dice}}$): NW-Dice enhances geometric sensitivity to edges by dynamically weighting pixels based on local discrepancies. The unified formulation combines spatial and class-wise weighting:
\begin{equation}
w_{i,j}\ =\ 1\ +\ \sum_{k=1}^{r} I \left[N_k(p_{i,j})\ \neq\ N_k(g_{i,j})\right]
\end{equation}
\begin{equation}
\mathcal{L}_{\text{NW-Dice}}\ =\ \frac{1}{C}\sum_{c=1}^{C}1 - \frac{2\sum(w^c \circ p^c \circ g^c)}{\sum(w^c \circ p^c) + \sum(w^c \circ g^c)}
\end{equation}
where $w_{i,j}$ is the weight of pixel $(i,j)$, $r$ is the maximum neighborhood radius (set to 5), $I$ is the indicator function (1 if different, 0 otherwise), $N_k$ is the $k$-th order neighborhood convolution operation with a kernel size of $(2k+1)\times(2k+1)$, $p_{i,j}$ is the predicted probability at $(i,j)$, $g_{i,j}$ is the ground truth label at $(i,j)$, $C$ denotes the number of classes, and $c$ represents the class weights.

\section{Experiments and Results}

\subsection{Dataset and implementations}

{\bf Dataset.} The baseline experiments have been carried out on data from the MICCAI 2023 Grand Challenge \cite{lu2022jnu}, which includes a total of 5,101 annotated images. To ensure a scientifically sound approach to model training, validation, and testing, the dataset was randomly divided into training, validation, and test sets at ratios of 70\%, 10\%, and 20\% respectively. For the sake of experimental simplicity, all training images underwent uniform preprocessing operations, including resizing and normalization. Each image has been paired with accurately delineated segmentation masks of the Fetal Head (FH) and Pubic Symphysis (PS), enabling effective development and evaluation of relevant segmentation models. For generalization experiments, this study used annotated images from the MICCAI 2024 Grand Challenge \cite{chen2024psfhs} which comprises a total of 300 images.

{\bf Implementation details.} In this study, we utilized the AoP-SAM \cite{zhou2025segment} model, a variant specifically designed for pubic symphysis-fetal head segmentation, and with the pre-trained weights it adopts. The model was implemented on Ubuntu 20.04 and utilized Python 3.8, PyTorch 1.10, and CUDA 11.3. The network training regimen encompassed 30,000 iterations, employing the Stochastic Gradient Descent (SGD) optimizer, configured with a momentum of 0.9 and a weight decay of 0.0001. We adopted a batch size 16, comprising an equal split of eight labeled and eight unlabeled images to support semi-supervised learning paradigms. The training commenced with an initial learning rate of 0.01, which was dynamically adjusted using a clustering-based learning rate strategy. Additionally, we introduced data noise perturbation within the range of [-0.2, 0.2] to enhance model robustness against input variability.

{\bf Comparison Strategy.} To verify the effectiveness of our DSTCS, We compared it against several state-of-the-art SSL approaches, including Mean Teacher (MT) \cite{tarvainen2017mean}, Interpolation Consistency (ICT) \cite{verma2022interpolation}, Uncertainty-Aware Mean Teacher (UAMT) \cite{yu2019uncertainty}, Deep Adversarial Network (DAN) \cite{zhang2017deep}, Deep Co-Training (DCT) \cite{qiao2018deep}, (CLB) \cite{liu2023semi}, cross pseudo-supervision (CPS) \cite{chen2021semi}, and Cross Teaching between CNN and Transformer (CTCT) \cite{luo2022semi}, Feature Similarity and Reliable Region Enhancement Model (FSRENet) [98], and Self-Ensembling Method with Consistency-Aware Pseudo-Labeling (S4CVnet) \cite{wang2022cnn}.

{\bf Evaluation Metrics.} The performance of these models was quantitatively evaluated using three established metrics: the Dice Similarity Coefficient (DSC), the 95\% Hausdorff Distance (HD95), and the Average Surface Distance (ASD). These metrics facilitate a thorough assessment of the segmentation models’ accuracy and consistency, critical for validating their clinical applicability.

\subsection{Comparison with other methods}
\begin{figure}[htb]
	\centering
	\includegraphics[width=1\linewidth]{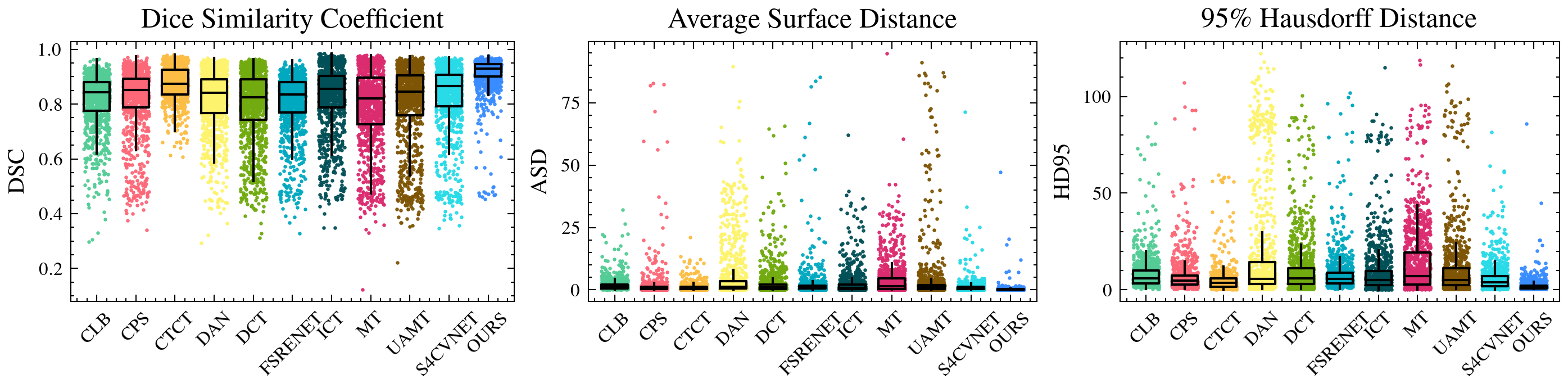}
	\caption{Box plots with overlaid scatter points comparing the performance distribution of different methods tested with 20\% labeled data.}
	\label{fig:box}
\end{figure}

\begin{figure}[htb]
	\centering
	\includegraphics[width=1\linewidth]{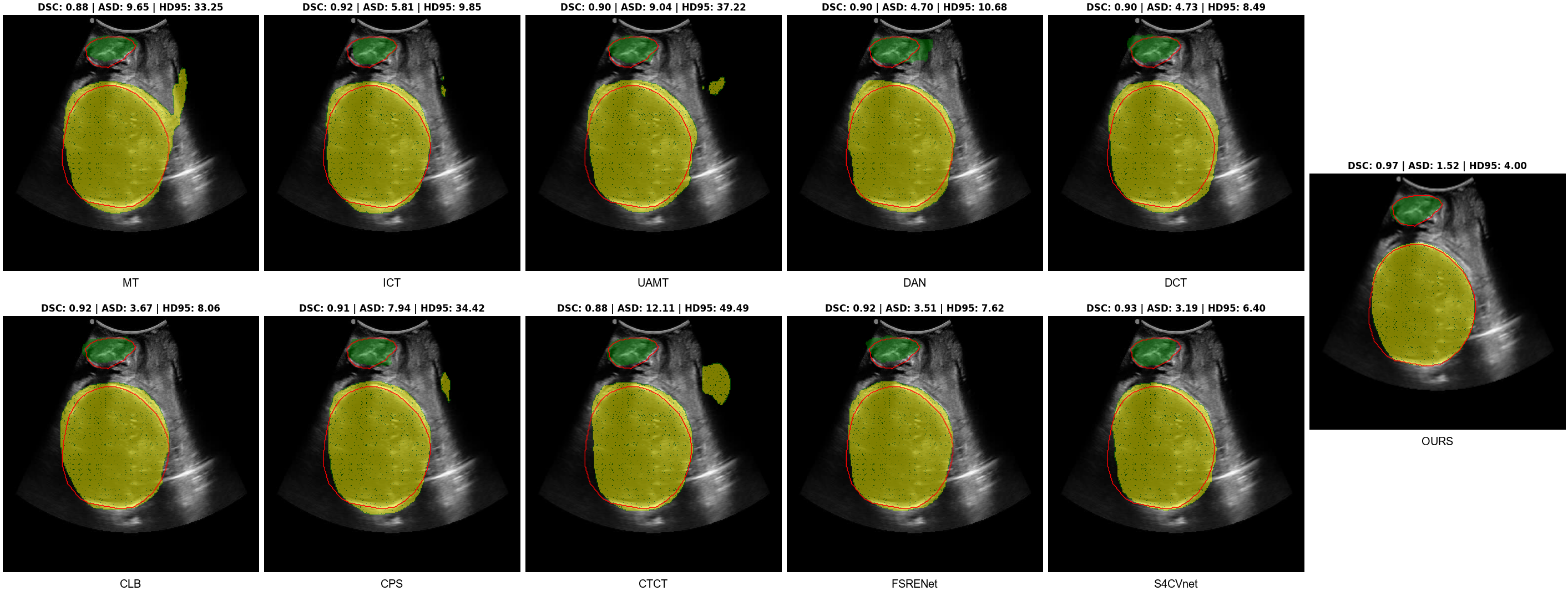}
	\caption{ Visual comparison of different methods using 20\% labeled data for testing.
		Red indicates ground truth annotations, green represents PS prediction results, and yellow denotes FH prediction results.}
	\label{fig:vis}
\end{figure}

Table \ref{tab:results1} demonstrates that, at 20\% labeled data ratios, DSTCS significantly outperformed all other competitors across all segmentation evaluation metrics. Notably, it achieved a PSFH-DSC of 0.911, far surpassing other methods (e.g., CTCT's 0.842), while its PSFH-HD95 is as low as 2.070, whereas other methods generally exceed 5.9. Further analysis from Fig \ref{fig:box} reveals that the performance fluctuation range of DSTCS is smaller, with significantly fewer outliers. Additionally, Fig \ref{fig:vis} clearly show that DSTCS outperformed the others in two key visual evaluation dimensions: boundary localization accuracy and regional overlap consistency.

\begin{table}[ht]
	\centering
	\caption{Quantitative comparison results on 20\% labeled data. The best result is in bold.}
	\label{tab:results1}
	\resizebox{0.48\textwidth}{!}{%
		\begin{tabular}{lccc}
			\toprule
			\multirow{2}{*}{Method} & \multicolumn{1}{c}{DSC(↑)} & \multicolumn{1}{c}{ASD(↓)} & \multicolumn{1}{c}{HD95(↓)}\\
			& \multicolumn{1}{c}{(PS/FH/PSFH)} & \multicolumn{1}{c}{(PS/FH/PSFH)} & \multicolumn{1}{c}{(PS/FH/PSFH)}\\
			\midrule
			MT & 0.701/0.874/0.787 & 5.350/2.841/4.096 & 15.566/13.107/14.337 \\
			ICT & 0.755/0.894/0.825 & $\underline{0.756}$/1.764/2.460 & 11.619/7.652/9.635 \\
			UAMT & 0.713/0.884/0.798 & 6.410/1.886/4.148 & 14.437/8.033/11.235 \\
			DAN & 0.742/0.882/0.812 & 9.763/1.843/5.803 & 28.006/8.564/18.285 \\
			DCT & 0.708/0.881/0.795 & 3.829/1.953/2.891 & 13.150/8.774/10.962 \\
			CLB & 0.751/0.874/0.813 & 2.380/1.498/1.939 & 9.641/6.934/8.287 \\
			CPS & 0.742/0.885/0.813 & 2.265/1.554/1.910 & 8.412/6.175/7.294 \\
			CTCT & $\underline{0.780}$/0.904/$\underline{0.842}$ & 1.542/1.567/1.554 & 7.035/7.353/7.194 \\
			FSRENet & 0.727/0.881/0.804 & 3.182/1.217/2.200 & 10.259/6.466/8.363 \\
			S4CVnet & 0.740/$\underline{0.905}$/0.823 & 1.378/$\underline{0.934}$/$\underline{1.156}$ & $\underline{6.834}$/$\underline{4.980}$/$\underline{5.907}$ \\
			DSTCS(ours) & \textbf{0.871/0.952/0.911} & \textbf{0.514/0.159/0.336} & \textbf{2.919/1.22/2.070} \\
			\bottomrule
		\end{tabular}
	}
\end{table}

\subsection{Generalization experiment}
As summarized in Table \ref{tab:results2}, gneralization experiments on the annotated dataset of MICCAI 2024 demonstrate that the proposed DSTCS achieved excellent performance in the PSFH segmentation task, with a DSC of 0.886, an ASD of 0.273 mm, and an HD95 of 3.975 mm. It is worth noting that it successfully segmented the pubic symphysis and fetal head structures in all test images in a complete and accurate manner, with no missed cases reported.

\begin{table}[ht]
	\centering
	\caption{Generalization Experiment results on 20\% labeled data. The best result is in bold.}
	\label{tab:results2}
	\small
	\resizebox{0.48\textwidth}{!}{
		\begin{tabular}{lcccr}
			\toprule
			\multirow{2}{*}{Method} & \multicolumn{1}{c}{DSC(↑)} & \multicolumn{1}{c}{ASD(↓)} & \multicolumn{1}{c}{HD95(↓)} & \multirow{2}{*}{Lossing(↓)} \\
			& \multicolumn{1}{c}{(PS/FH/PSFH)} & \multicolumn{1}{c}{(PS/FH/PSFH)} & \multicolumn{1}{c}{(PS/FH/PSFH)} & \\
			\midrule
			MT & 0.626/0.897/0.761 & 2.334/1.256/1.795 & 14.258/7.016/10.637 & 2 \\
			ICT & 0.644/0.898/0.771 & 2.392/1.181/1.786 & 14.012/6.855/10.433 & 5 \\
			UAMT & 0.580/0.889/0.734 & 2.869/1.047/1.958 & 14.349/7.153/10.751 & 17 \\
			DAN & $\underline{0.722}$/0.891/$\underline{0.807}$ & 5.167/1.141/3.154 & 18.561/7.406/12.984 & 2 \\
			DCT & 0.625/0.892/0.759 & 2.150/1.216/1.683 & 13.960/7.121/10.540 & $\underline{1}$ \\
			CLB & 0.703/$\underline{0.903}$/0.803 & 3.698/0.785/2.242 & 14.158/$\underline{5.691}$/9.924 & 2 \\
			CPS & 0.505/0.902/0.703 & 2.275/0.960/1.618 & 13.979/6.395/10.187 & 38 \\
			CTCT & 0.654/0.893/0.774 & 1.460/1.207/1.333 & $\underline{11.195}$/8.104/9.650 & 3 \\
			FSRENet & 0.647/0.894/0.771 & $\underline{1.026}$/0.745/$\underline{0.886}$ & 11.495/6.586/$\underline{9.040}$ & 4 \\
			S4CVnet & 0.604/0.899/0.751 & 2.171/$\underline{0.683}$/1.427 & 12.396/6.863/9.630 & 12 \\
			DSTCS(ours) & \textbf{0.851/0.921/0.886} & \textbf{0.326/0.220/0.273} & \textbf{3.872/4.078/3.975} & \textbf{0} \\
			\bottomrule
		\end{tabular}
	}
\end{table}

\subsection{Ablation study}

{\bf Comparative Analysis of Model Configurations.} We investigated the impact of different backbone models on the framework's performance, and the results are summarized in Table \ref{tab:results3}. It demonstrates that the adoption of AoP-SAM as the backbone has achieved the best segmentation performance, which not only enhanced single-target segmentation quality but also promoted collaborative learning among multiple structures. This approach maintains high overlap accuracy while significantly improving boundary precision.

\begin{table}[ht]
	\centering
	\caption{Performance comparison of different model configurations in experiments.}
	\label{tab:results3}
	\small
	\resizebox{0.48\textwidth}{!}{
		\begin{tabular}{lllccc}
			\toprule
			\multirow{2}{*}{S1} & \multirow{2}{*}{S2} & \multirow{2}{*}{T} & \multicolumn{1}{c}{DSC(↑)} & \multicolumn{1}{c}{ASD(↓)} & \multicolumn{1}{c}{HD95(↓)} \\
			& & & \multicolumn{1}{c}{(PS/FH/PSFH)} & \multicolumn{1}{c}{(PS/FH/PSFH)} & \multicolumn{1}{c}{(PS/FH/PSFH)} \\
			\midrule
			UNet & SUNet & SUNet & 0.837/0.932/0.884 & 0.598/0.429/0.514 & 3.734/$\underline{2.295}$/3.014 \\
			SUNet & UNet & UNet & 0.845/0.890/0.867 & 0.902/1.515/1.208 & 4.205/6.305/5.255 \\
			UNet & UNet & UNet & 0.851/0.900/0.876 & 0.781/1.401/1.091 & 3.957/6.394/5.175 \\
			SUNet & SUNet & SUNet & 0.820/0.920/0.870 & 0.966/$\underline{0.433}$/0.699 & 4.336/2.810/3.573 \\
			AoP-SAM & UNet & UNet & \textbf{0.896/0.943/0.920} & \textbf{0.294/0.272/0.284} & \textbf{1.961/1.624/1.793} \\
			UNet & AoP-SAM & AoP-SAM & $\underline{0.884}$/\underline{0.929}/\underline{0.906} & $\underline{0.333}$/0.498/$\underline{0.415}$ & $\underline{2.241}$/2.532/$\underline{2.387}$ \\
			\bottomrule
		\end{tabular}
	}
\end{table}

{\bf Discussion on the Parameters of NW-Dice Loss.} As detailed in Table \ref{tab:hyperparam}, this study analyzes the impact of class weight (class\_weight) and local neighborhood range (r) in the Neighborhood Weighted Dice Loss. Experiments demonstrated that class\_weight = [1.0, 2.0, 1.0] combined with r = 5 achieved the highest DICE (0.927), the lowest ASD (0.181), and a relatively low HD95 (1.439). Appropriately increasing the weight for PS improved its segmentation performance, but excessively high weights (e.g., 3.0) lead to performance degradation. A local range of r = 5 effectively captures contextual information, while values that are too large or too small resulted in performance deterioration.

\begin{table}[ht]
	\centering
	\caption{Hyperparameter analysis of class weights and local range in NW-Dice loss}
	\label{tab:hyperparam}
	\resizebox{0.48\textwidth}{!}{
		\begin{tabular}{lccc}
			\toprule
			\multirow{2}{*}{Hyperparameter Settings} & \multicolumn{1}{c}{DSC(↑)} & \multicolumn{1}{c}{ASD(↓)} & \multicolumn{1}{c}{HD95(↓)} \\
			& \multicolumn{1}{c}{(PS/FH/PSFH)} & \multicolumn{1}{c}{(PS/FH/PSFH)} & \multicolumn{1}{c}{(PS/FH/PSFH)} \\
			\midrule
			class\_weight=[1.0,1.5,1.0], r=5 & 0.898/$\bf{0.955}$/0.926 & 0.251/0.132/0.191 & 1.886/$\bf{0.989}$/$\bf{1.438}$ \\
			class\_weight=[1.0,1.0,1.0], r=5 & 0.899/0.954/0.927 & 0.255/0.143/0.199 & 1.853/1.034/1.443 \\
			class\_weight=[1.0,3.0,1.0], r=5 & 0.895/0.954/0.925 & 0.360/0.148/0.254 & 2.150/1.016/1.583 \\
			class\_weight=[1.0,1.5,1.0], r=7 & 0.896/0.952/0.924 & 0.340/0.176/0.258 & 2.126/1.108/1.617 \\
			class\_weight=[0.5,1.5,1.0], r=7 & 0.898/0.954/0.926 & 0.247/0.128/0.187 & 1.904/1.021/1.463 \\
			class\_weight=[1.0,2.0,1.0], r=5 & $\bf{0.900}$/0.954/$\bf{0.927}$ & $\bf{0.245}$/$\bf{0.118}$/$\bf{0.181}$ & $\bf{1.843}$/1.034/1.439 \\
			class\_weight=[1.0,2.0,1.0], r=3 & 0.897/$\bf{0.955}$/0.926 & 0.250/0.160/0.205 & 1.909/1.023/1.466 \\
			class\_weight=[1.0,2.0,1.0], r=7 & 0.898/$\bf{0.955}$/0.926 & 0.258/0.154/0.206 & 1.898/0.990/1.444 \\
			class\_weight=[0.5,1.5,1.0], r=7 & 0.898/0.954/0.926 & 0.247/0.128/0.187 & 1.904/1.021/1.463 \\
			\bottomrule
		\end{tabular}
	}
\end{table}

\subsection{Impact of annotation ratio on segmentation performance}
\begin{figure}[htb]
	\centering
	\includegraphics[width=1\linewidth]{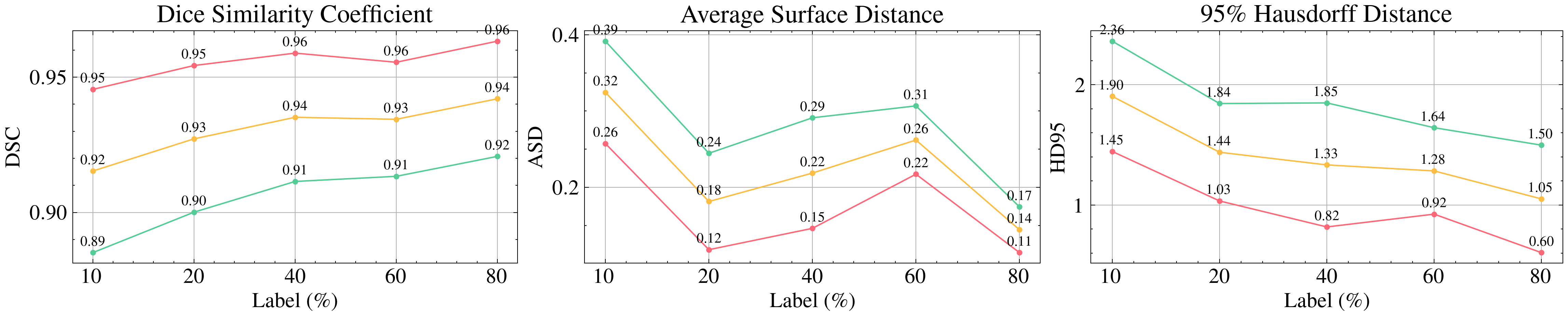}
	\caption{Segmentation performance versus annotation ratios. Red, green and yellow curves denote the FH, PS, and PSFH, respectively.}
	\label{fig:bai}
\end{figure}

Fig \ref{fig:bai} demonstrates that segmentation performance improved with more annotated data. Notably, when the annotation ratio exceeded 40\%, the improvement in PS-DSC began to plateau, indicating that the model’s recognition of the main structure is nearing saturation.
		
\section{Conclusion}

This paper proposes DSTCS, a novel semi-supervised framework, to address key challenges in pubic symphysis–fetal head (PSFH) ultrasound image segmentation. Furthermore, the proposed In-situ Edge Patch Overlay (IEPO) augmentation strategy and Neighborhood Weighted Dice (NWD) Loss effectively improve boundary modeling and spatial sensitivity. Experiments on MICCAI 2023 and 2024 PSFH benchmark datasets demonstrate that it significantly outperforms existing methods in both segmentation accuracy (DSC) and boundary metrics (ASD, HD95), exhibiting stronger robustness in high-noise and low-contrast scenarios.

\bibliographystyle{IEEEbib}

\end{document}